\documentclass{article}
\usepackage{spconf,amsmath,graphicx,hyperref}

\graphicspath{ {./imgs/} } 
\usepackage{multirow} 

\title{Enhancing Document-Level Machine Translation via Filtered Synthetic Corpora and Two-Stage LLM Adaptation}
\name{Ireh Kim$^{1}$, Tesia Sker$^{1}$, Chanwoo Kim$^{1\dagger}$}
\address{$^{1}$ Department of Artificial Intelligence \\
         Korea University, Seoul, South Korea \\}

\begin{document}
%
\maketitle
\begin{abstract}
In Machine Translation, Large Language Models (LLMs) have generally underperformed compared to conventional encoder–decoder systems and thus see limited adoption. 
However, LLMs excel at modeling contextual information, making them a natural fit for document-level translation tasks where coherence across sentences is crucial.
Despite this potential, document-level MT with LLMs faces two key challenges: (1) the scarcity of large-scale, high-quality document-level parallel data; and (2) the propensity of LLMs to introduce hallucinations and omissions during generation. 
To address these challenges, we propose a two-stage fine-tuning strategy leveraging LLM-augmented document-level data.
First, we augment data by converting summarization data into document-level parallel data using a LLM, and then filter it using multiple metrics, leveraging sacreBLEU, COMET, and LaBSE-based cosine similarity—to improve data quality.
Finally, we employ a two-stage fine-tuning strategy: first fine-tuning on the abundant sentence‑level MT resources, and then on the filtered document-level corpus.
\end{abstract}

\section{Introduction}
\label{sec:intro}


Large language models (LLMs) have recently shown promising potential in machine translation (MT), including context modeling and fluency.
However, LLMs are prone to hallucinations and omissions during translation~\cite{bang-etal-2023-multitask}, raising concerns regarding the translation reliability.
As a result, conventional MT models are still predominantly used in practical machine translation applications.

In Machine Translation, hallucinations, which refer to the inclusion of information in the translation that does not appear in the source text, and omissions, which refer to the exclusion of information in the translation that appears in the source text, are not captured by traditional metrics such as sacreBLEU~\cite{post-2018-call}. 
To address these issues, various approaches have been proposed. 
Bag-of-vectors sentence similarity (BVSS)~\cite{martindale-etal-2019-identifying} detects hallucinations and omissions through statistical clustering, 
while xCOMET~\cite{guerreiro-etal-2024-xcomet} extends COMET~\cite{rei-etal-2020-comet} to jointly perform sentence-level evaluation and error span prediction. 
More recently, the Word Alignment Preference (WAP) framework~\cite{wu-etal-2024-word} incorporates coverage-based signals into preference optimization, further improving robustness against these errors.

To achieve more natural translations and improve translation quality, it is essential to leverage LLMs for document-level translation, enabling more effective utilization of contextual information while minimizing hallucinations and omissions. 
However, existing MT datasets such as News Commentary and Europarl~\cite{TIEDEMANN12.463} are primarily designed for sentence-level translation, making it challenging for LLMs to fully reflect contextual information.

In this work, we propose a two-stage fine-tuning strategy leveraging LLM-augmented document-level data.
First, we generate English-German translation pairs from the CNN/Daily Mail summarization dataset~\cite{see-etal-2017-get} using the \texttt{Llama\hspace{0pt}3.1-8B-Instruct} model~\cite{grattafiori2024llama} and construct pseudo-references using Google Translate.
Then, we apply multi-metric filtering—using sacreBLEU, COMET, and cosine similarity based on the \texttt{LaBSE} model~\cite{feng-etal-2022-language}—to mitigate hallucinations and omissions in the augmented data.
This cosine similarity method, which has been shown to be effective in detecting hallucinations~\cite{dale-etal-2023-detecting}, is referred to as LaBSE-CosSim throughout the remainder of this paper.
Finally, we fine-tune the \texttt{Llama-3.2\hspace{0pt}-1B-Instruct} model~\cite{grattafiori2024llama} using a two-stage approach, first on sentence-level data and then on the filtered document-level corpus.

\begin{table*}[h!]
    \centering
    \caption{Sizes of filtered datasets based on different thresholds.}
    \small
    \begin{tabular}{l|c|ccccccc}
    \hline
    \begin{tabular}[c]{@{}l@{}}Thresholds/\\ Filtering Methods\end{tabular}               & \multicolumn{1}{l|}{\begin{tabular}[c]{@{}l@{}}No\\Filtering\end{tabular}}  & \multicolumn{1}{c}{sacreBLEU}                             & \multicolumn{1}{c}{COMET}                                 & \multicolumn{1}{c}{\begin{tabular}[c]{@{}l@{}}LaBSE-\\CosSim\end{tabular}}  & \multicolumn{1}{l}{\begin{tabular}[c]{@{}l@{}}sacreBLEU \&\\ COMET\end{tabular}}  & \multicolumn{1}{l}{\begin{tabular}[c]{@{}l@{}}sacreBLEU \&\\ LaBSE-CosSim\end{tabular}}  & \multicolumn{1}{l}{\begin{tabular}[c]{@{}l@{}}COMET \&\\ LaBSE-CosSim\end{tabular}}  & \multicolumn{1}{c}{All}                                   \\ \hline
    \begin{tabular}[c]{@{}l@{}}sacreBLEU 30\\ COMET 0.7\\ LaBSE-CosSim 0.8\end{tabular}   &                                                                             & \begin{tabular}[c]{@{}c@{}}18,439\\(92.2\%)\end{tabular}  & \begin{tabular}[c]{@{}c@{}}19,925\\(99.6\%)\end{tabular}  & \begin{tabular}[c]{@{}c@{}}19,917\\(99.6\%)\end{tabular}                    & \begin{tabular}[c]{@{}c@{}}18,430\\(92.2\%)\end{tabular}                          & \begin{tabular}[c]{@{}c@{}}18,420\\(92.1\%)\end{tabular}                                 & \begin{tabular}[c]{@{}c@{}}19,868\\(99.3\%)\end{tabular}                             & \begin{tabular}[c]{@{}c@{}}18,411\\(92.1\%)\end{tabular}  \\ \cline{1-1} \cline{3-9} 
    \begin{tabular}[c]{@{}l@{}}sacreBLEU 35\\ COMET 0.75\\ LaBSE-CosSim 0.85\end{tabular} & \begin{tabular}[c]{@{}c@{}}20,000\\(100\%)\end{tabular}                     & \begin{tabular}[c]{@{}c@{}}15,139\\(75.7\%)\end{tabular}  & \begin{tabular}[c]{@{}c@{}}19,526\\(97.8\%)\end{tabular}  & \begin{tabular}[c]{@{}c@{}}19,554\\(97.8\%)\end{tabular}                    & \begin{tabular}[c]{@{}c@{}}15,016\\(75.0\%)\end{tabular}                          & \begin{tabular}[c]{@{}c@{}}14,913\\(74.6\%)\end{tabular}                                 & \begin{tabular}[c]{@{}c@{}}19,137\\(95.7\%)\end{tabular}                             & \begin{tabular}[c]{@{}c@{}}14,791\\(74.0\%)\end{tabular}  \\ \cline{1-1} \cline{3-9} 
    \begin{tabular}[c]{@{}l@{}}sacreBLEU 40\\ COMET 0.80\\ LaBSE-CosSim 0.90\end{tabular} &                                                                             & \begin{tabular}[c]{@{}c@{}}8,914\\(44.6\%)\end{tabular}   & \begin{tabular}[c]{@{}c@{}}16,471\\(82.4\%)\end{tabular}  & \begin{tabular}[c]{@{}c@{}}14,220\\(71.1\%)\end{tabular}                     & \begin{tabular}[c]{@{}c@{}}8,234\\(41.2\%)\end{tabular}                           & \begin{tabular}[c]{@{}c@{}}6,633\\(33.2\%)\end{tabular}                                 & \begin{tabular}[c]{@{}c@{}}11,447\\(57.2\%)\end{tabular}                             & \begin{tabular}[c]{@{}c@{}}6,042\\(30.2\%)\end{tabular}  \\ \hline
    \end{tabular}
    \label{tab:data}
\end{table*}

\vspace{-1.0cm}
\section{Proposed Method}
\label{sec:ProposedMethod}
\vspace{-0.5cm}


To achieve natural translations, MT must move beyond sentence-level processing and incorporate broader context. 
LLMs are well-suited for this but often suffer from hallucinations and omissions, and available document-level En→De datasets remain limited (e.g., 10,639 pairs in News Commentary v16, 9,202 in Europarl v7, with only ~3,000 valid pairs). 
This scarcity hinders effective fine-tuning. 
To address these issues, we propose a document-level data augmentation pipeline and a two-stage fine-tuning strategy: first training on sentence-level data, 
then adapting to document-level data with mechanisms to reduce hallucinations and omissions.

\subsection{Document-level MT data augmentation} 
When fine-tuning a Large Language Model (LLM) for Machine Translation (MT) tasks, to obtain a sufficient amount of document-level MT data, reflecting contextual information, 
we follow a two-step process: (1) augmentation using a LLM, and (2) filtering with multiple evaluation metrics to construct high-quality data, mitigating hallucinations and omissions.


\begin{figure*}[h!]
    \centerline{\includegraphics[width=0.8\linewidth]{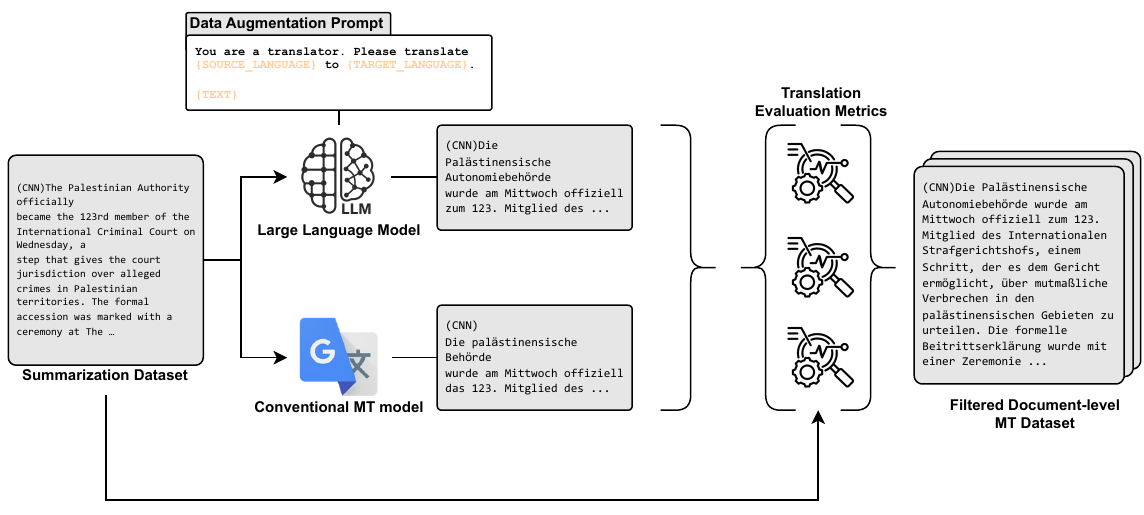}}
    \caption{Overview of the document-level MT data augmentation and filtering pipeline. We convert a summarization dataset (CNN/Daily Mail) into document-level MT pairs using a large language model (\texttt{Llama-3.1-8B-Instruct}), then apply filtering using sacreBLEU, COMET, and LaBSE-CosSim. For reference-based metrics, pseudo references are generated by a conventional MT model (Google Translate) known to produce fewer hallucinations and omissions.}
    \label{fig:flow}
\end{figure*}

\hfill
\newline \noindent\textbf{Augmentation via Prompted LLM Translation.} For data augmentation, we utilize a larger LLM than the target fine-tuning model.
While this process allows us to obtain document-level MT data that reflect contextual information, LLMs can introduce hallucinations and omissions during the data augmentation process, degrading data quality and potentially leading to performance deterioration in the fine-tuned model due to the propagation of such errors.

\hfill
\newline \noindent\textbf{Filtering Methods.} To address this issue, we employ three evaluation metrics—sacreBLEU, COMET, and LaBSE-CosSim—to filter the augmented data.
sacreBLEU measures n-gram overlap with a reference, while COMET is a reference-based learned metric which predict human MT quality judgments. 
In contrast, LaBSE-CosSim directly evaluates semantic similarity without references.
To handle length constraint of \texttt{LaBSE} model, we compute sentence-level embeddings for source and translation, average them, and calculate cosine similarity.

As both sacreBLEU and COMET rely on reference translations, we generate pseudo references using Google Translate—a conventional sentence-level MT system known for producing reliable translations with fewer hallucinations. 
These references are then paired with the LLM outputs, and we compute the corresponding scores. 
Based on empirically determined thresholds for each metric, we discard low-quality examples.
The filtering statistics are summarized in Table~\ref{tab:data}.

We also considered using xCOMET and Doc-COMET~\cite{vernikos-etal-2022-embarrassingly}, but found that these metrics are not suitable in our setting.
Although xCOMET is capable of evaluating hallucinations, it does not support document-level inputs and thus fails to operate properly in our setting.
While Doc-COMET supports document-level evaluation, it requires sentence-level alignment between the source and the translation, which is not available for our augmented data.
The overall augmentation and filtering process is illustrated in Fig~\ref{fig:flow}.

\subsection{Two-stage LLM fine-tuning}
We propose a two-stage fine-tuning strategy to enhance its performance in document-level machine translation (MT). 
In the first stage, the model is fine-tuned on a sentence-level MT dataset to facilitate more effective training on document-level MT data in the subsequent stage.
We utilize conventional sentence-level MT datasets such as News Commentary and Europarl, which are more abundant and offer high-quality parallel sentence pairs.

In the second stage, the model is further fine-tuned on a filtered, LLM-augmented document-level dataset to incorporate contextual information more effectively. 
This step enables the model to better incorporate contextual information across sentences, which is essential for coherent document-level translation. 
By leveraging both abundant sentence-level data and context-rich augmented document-level data, this two-stage strategy improves the fluency and consistency of the translated output.

\vspace{-0.4cm}
\section{Experiments}
\label{sec:experiments}
\vspace{-0.4cm}

\subsection{Datasets}
\textbf{Augmented document-level dataset.} For document-level MT data augmentation, we adopted an approach that converts the CNN/Daily Mail summarization dataset into a document-level translation dataset, leveraging its coherent narrative structure and formal jornalistic style.
Unlike subtitle datasets that rely on visual or auditory context, CNN/Daily Mail provides sufficient textual context for MT. 
For the efficient data augmentation, we randomly sampled 20K documents (7\% of 287K instances), yielding 17.4M English and 25.4M German tokens.
Note that we removed automatically generated prefixes (e.g., “Hier ist die Übersetzung ins Deutsche:”) inserted by the Llama-3.1-8B-Instruct model during augmentation process.

\vspace{-0.2cm}
\hfill
\newline \noindent\textbf{Human-labeled dataset.} 
For the first stage of fine-tuning, we used the En–De portion of News Commentary v16, sampling 20K sentence pairs from 399K instances (8.2M English and 13.4M German tokens) to match the domain of the augmented dataset.

\subsection{Experiments Details}

For document-level machine translation, we target English-to-German (En→De) translation and select the well-known open-source model \texttt{Llama-3.2\hspace{0pt}-1B-Instruct} as our base model, balancing contextual ability with manageable size.
\texttt{Llama-3.2\hspace{0pt}-1B-Instruct} is sufficiently pre-trained on both English and German, making it suitable for basic translation tasks.
To generate the augmented document-level MT dataset, we employ a relatively larger model, \texttt{Llama-3.1-8B-Instruct}, which performs better on long-context translation.

To demonstrate the effectiveness of our proposed two-stage fine-tuning strategy and filtering methods, 
we randomly sample 2,000 instances from the augmented dataset and compare the performance of the following two models:
\begin{enumerate}
    \item A baseline model trained solely on the document-level augmented data without sentence-level pretraining.
    \item A model fine-tuned using the proposed two-stage strategy—first trained on sentence-level data, then further fine-tuned on the same augmented document-level data.
\end{enumerate}
To assess the impact of filtering, we use filtered datasets retaining at least 70\% of the original 20K augmented instances, sample 2,000 from each, and train models with the two-stage strategy.
The sentence-level stage uses 20K News Commentary v16 pairs, and evaluation is performed on 1,500 document-level instances with sacreBLEU, COMET, LaBSE-CosSim, and their geometric mean.

\vspace{-0.4cm}
\begin{table}[h!]
    \centering
    \caption{Comparison between models trained only on unfiltered augmented data and those trained using the proposed two-stage fine-tuning strategy.}
    \begin{tabular}{l||c|c}
    \hline
    Fine-tuning Strategy & \begin{tabular}[c]{@{}c@{}}Document-level\\only\end{tabular} & \begin{tabular}[c]{@{}c@{}}Two-stage\\fine-tuning\end{tabular}          \\ \hline
    sacreBLEU         & 11.24                                                        & \bf{15.07}                                                              \\
    COMET             & 0.618                                                        & \bf{0.697}                                                              \\
    LaBSE-CosSim      & 0.764                                                        & \bf{0.852}                                                              \\ \hline
    Geometric Mean    & 1.744                                                        & \bf{2.076}                                                              \\ \hline
    \end{tabular}
    \label{tab:baselines_results}
\end{table}

\begin{table*}[h!]
    \centering
    \small
    \caption{Evaluation results of models fine-tuned on data filtered using a single method. Each model is trained using the two-stage fine-tuning strategy with datasets filtered by a single metric.}
    \begin{tabular}{l||c|cc|ccc|ccc}
    \hline
    Filtering Methods & \begin{tabular}[c]{@{}c@{}}No Filtering\\ (Augmented data)\end{tabular} & \multicolumn{2}{c|}{sacreBLEU} & \multicolumn{3}{c|}{COMET}           & \multicolumn{3}{c}{LaBSE-CosSim}   \\ \hline
    Thresholds        & -                                                                       & 30             & 35            & 0.7     & 0.75         & 0.8         & 0.8       & 0.85      & 0.9        \\ \hline
    sacreBLEU         & 15.07                                                                   & 15.58          & \bf{15.85}    & 15.66   & 13.8         & 15.53       & 14.04     & 15.53     & 15.74      \\ 
    COMET             & 0.697                                                                   & 0.698          & 0.696         & 0.697   & \bf{0.700}   & \bf{0.700}  & 0.691     & 0.694     & 0.698      \\ 
    LaBSE-CosSim      & 0.852                                                                   & 0.851          & 0.855         & 0.847   & 0.848        & 0.854       & 0.843     & 0.850     & \bf{0.856} \\ \hline
    Geometric Mean    & 2.076                                                                   & 2.100          & \bf{2.113}    & 2.099   & 2.016        & 2.102       & 2.013     & 2.092     & 2.111      \\ \hline
    \end{tabular}
    \label{tab:doc_single_results}
\end{table*}

\begin{table*}[h!]
    \centering
    \small
    \caption{Evaluation results of models trained on datasets filtered using two combined metrics. Models are trained with two-stage fine-tuning on datasets filtered by metric combinations.}
    \begin{tabular}{l||c|cc|cc|ccc}
    \hline
    Filtering Methods & \begin{tabular}[c]{@{}c@{}}No Filtering\\ (Augmented data)\end{tabular} & \multicolumn{2}{c|}{sacreBLEU \& COMET} & \multicolumn{2}{c|}{sacreBLEU \& LaBSE-CosSim} & \multicolumn{3}{c|}{COMET \& LaBSE-CosSim} \\ \hline
    Thresholds        & -                                                                       & 30 \& 0.7          & 35 \& 0.75         & 30 \& 0.8             & 35 \& 0.85             & 0.7 \& 0.8  & 0.75 \& 0.85  & 0.8 \& 0.9   \\ \hline
    sacreBLEU         & 15.07                                                                   & 15.82              & \bf{15.99}         & 15.79                 & 15.89                  & 15.57       & 15.43         & 15.65                        \\
    COMET             & 0.697                                                                   & 0.700              & 0.699              & 0.698                 & \bf{0.701}             & 0.697       & 0.697         & 0.700        \\
    LaBSE-CosSim      & 0.852                                                                   & 0.855              & 0.856              & 0.857                 & 0.855                  & 0.851       & \bf{0.858}    & 0.851        \\ \hline
    Geometric Mean    & 2.076                                                                   & 2.116              & \bf{2.123}         & 2.114                 & 2.120                  & 2.098       & 2.097         & 2.105        \\ \hline
    \end{tabular}
    \label{tab:doc_double_results}
\end{table*}

\begin{table*}[h!]
    \centering
    \small
    \caption{Evaluation results of models trained on datasets filtered using all three metrics.}
    \begin{tabular}{l||c|cc}
    \hline
    Filtering Methods & \begin{tabular}[c]{@{}c@{}}No Filtering\\ (Augmented data)\end{tabular} & \multicolumn{2}{c}{\begin{tabular}[c]{@{}c@{}}sacreBLEU \& COMET \& \\ LaBSE-CosSim\end{tabular}} \\ \hline
    Thresholds        & -                                                                       & 30 \& 0.7 \& 0.8           & 35 \& 0.75 \& 0.85                                                   \\ \hline
    sacreBLEU         & 15.07                                                                   & 15.90                      & \bf{15.96}                                                           \\
    COMET             & 0.697                                                                   & 0.699                      & \bf{0.701}                                                           \\
    LaBSE-CosSim      & 0.852                                                                   & 0.856                      & \bf{0.860}                                                           \\ \hline
    Geometric Mean    & 2.076                                                                   & 2.119                      & \bf{2.127}                                                           \\ \hline
    \end{tabular}
    \label{tab:doc_triple_results}
\end{table*}

\vspace{-0.8cm}
\section{Results}
\label{sec:results}
\vspace{-0.3cm}


First, Table~\ref{tab:baselines_results} shows the performance of \texttt{Llama-3.2-1B-\hspace{0pt}Instruct} trained on 2,000 random instances from the unfiltered LLM-augmented dataset.
The results demonstrate that the proposed two-stage fine-tuning strategy—initial training on sentence-level data followed by fine-tuning on document-level data—consistently outperforms document-level-only training across all metrics: sacreBLEU, COMET, LaBSE-CosSim, and their geometric mean.

We further evaluate filtered datasets using single-metric thresholds presented in Table~\ref{tab:doc_single_results}. 
Models trained on filtered data generally surpass those trained on unfiltered data, with higher thresholds yielding better performance. 
Among these, sacreBLEU-based filtering achieves the most notable improvement.

Tables~\ref{tab:doc_double_results} and~\ref{tab:doc_triple_results} show results for datasets filtered using two or more metrics. 
Consistent with earlier observations, higher thresholds improve performance. 
Among two-metric combinations, sacreBLEU combined with COMET or LaBSE-CosSim yields stronger improvements than the COMET and LaBSE-CosSim pair. 
This indicates that combining metrics with different characteristics, such as lexical overlap captured by sacreBLEU and semantic evaluation captured by COMET or LaBSE-CosSim, provides complementary filtering effects.

Ultimately, the model trained on the dataset filtered using all three metrics achieves the best overall performance, with COMET 0.701, LaBSE-CosSim 0.860, and sacreBLEU 15.96. 
Although its sacreBLEU is slightly lower than the sacreBLEU and COMET combination, it shows balanced strength across all metrics.
Therefore, Applying the two-stage fine-tuning strategy on data filtered with sacreBLEU $\ge$ 35, COMET $\ge$ 0.75, and LaBSE-CosSim $\ge$ 0.85 is the most effective configuration.

\vspace{-0.6cm}
\section{Conclusion}
\label{sec:conclusion}
\vspace{-0.3cm}

In this work, we tackle key limitations of LLMs in document-level machine translation, including limited high-quality data and frequent hallucinations. 
We propose a framework combining LLM-based synthetic data augmentation, multi-metric filtering, and a two-stage fine-tuning strategy integrating sentence- and document-level learning.
Experiments on English-to-German translation show that 
(1) two-stage fine-tuning consistently improves document-level performance, 
(2) multi-metric filtering of augmented data significantly enhances quality, and 
(3) sacreBLEU works best in combination with COMET and LaBSE-CosSim, with optimal thresholds of sacreBLEU $\ge$ 35, COMET $\ge$ 0.75, and LaBSE-CosSim $\ge$ 0.85.
These results demonstrate the practical potential of LLMs for document-level MT under resource constraints and suggest that our approach can generalize to other domains and language pairs.

\vspace{-0.4cm}
\section{Acknowledgements}
\label{sec:acknowledgements}
\vspace{-0.3cm}
This work was supported by the IITP (Institute of Information Communications Technology Planning Evaluation)–ITRC (Information Technology Research Center) grant funded by the Korea government (Ministry of Science and ICT) (IITP2025-RS-2024-00436857), 
the IITP grant funded by the Korea government (MSIT) (No. RS-2019-II190079, Artificial Intelligence Graduate School Program, Korea University), 
the IITP under the Artificial Intelligence Star Fellowship support program to nurture the best talents (IITP2025-RS-2025-02304828) grant funded by the Korea government (MSIT), 
the IITP grant funded by the Korea government (MSIT) (No. RS-2025-25442867), 
and the National Research Foundation of Korea grant(NRF) funded by the Korea government (MSIT) (No. RS-2025-24535409).

\vspace{-0.4cm}
\bibliographystyle{IEEEbib}
\bibliography{strings,refs}

\end{document}